\documentclass[runningheads]{llncs}
\usepackage[T1]{fontenc}
\usepackage{graphicx}
\usepackage[dvipsnames]{xcolor}

\usepackage{array}
\usepackage{caption}
\usepackage{makecell}
\setcellgapes{1pt}   
\makegapedcells

\usepackage{subcaption}
\usepackage{tikz}
\usetikzlibrary{shapes.geometric, arrows, positioning, decorations.pathreplacing,calc}

\usepackage{amsmath}
\usepackage{amsfonts}
\usepackage[hidelinks]{hyperref}
\usepackage[ruled,linesnumbered,lined]{algorithm2e}

\usepackage{microtype}

\usepackage{multirow}
\begin{document}
\title{Socio-cognitive agent-oriented evolutionary algorithm with trust-based optimization}

%
\titlerunning{Socio-cognitive agent-oriented evolutionary algorithm\ldots}
%
\author{Aleksandra Urbańczyk\inst{1} \orcidID{0000-0002-6040-554X} \and
Krzysztof Czech\inst{1}
\and Piotr Urbańczyk\inst{2,1}\orcidID{0000-0001-8838-2354}
\and Marek Kisiel-Dorohinicki\inst{1}\orcidID{0000-0002-8459-1877}
\and Aleksander Byrski\inst{1}\orcidID{0000-0001-6317-7012}}
\authorrunning{A. Urbańczyk et al.}
%
\institute{AGH University of Krakow, Al. Mickiewicza 30, 30-059 Krakow, Poland
\email{\{aurbanczyk,purbanczyk,doroh,olekb\}@agh.edu.pl}, \email{krzysztof.czechh@gmail.com} \and
Jagiellonian University, ul. Gołębia 24, 31-007 Kraków, Poland
\email{piotr.urbanczyk@uj.edu.pl}}
\maketitle              
\begin{abstract}

This paper introduces the Trust-Based Optimization (TBO), a novel extension of the island model in evolutionary computation that replaces conventional periodic migrations with a flexible, agent-driven interaction mechanism based on trust or reputation.
 Experimental results demonstrate that TBO generally outperforms the standard island model evolutionary algorithm across various optimization problems. Nevertheless, algorithm performance varies depending on the problem type, with certain configurations being more effective for specific landscapes or dimensions. The findings suggest that trust and reputation mechanisms provide a flexible and adaptive approach to evolutionary optimization, improving solution quality in many cases.

\end{abstract}
\section{Introduction}\label{sec:intro}

\enlargethispage{2.5\baselineskip}

Evolutionary algorithms (EAs) have been widely used for solving complex optimization problems, with island models (IMs) enhancing scalability and maintaining population diversity (see, e.g. \cite{Saib_Abdessemed_Hocin_2024,Stanczak_2024}). However, these traditional approaches face several challenges, including premature convergence, inefficient information exchange, and imbalanced exploration-exploitation dynamics. The effectiveness of an island model largely depends on the communication strategy between subpopulations, where either excessive or insufficient migration can negatively impact performance. For instance, excessive communication reduces diversity among individuals across island populations, leading to rapid convergence. In contrast, insufficient communication prevents the exchange of crucial information between islands, causing them to evolve independently and failing to leverage the benefits of the island model.

The goal of developing a new relationship-based interaction mechanism is to address these typical issues of the island model. Unlike standard IMs, where migration serves as the primary means of interaction, our approach introduces an adaptive, agent-based mechanism that enables islands to exchange information based on trust or reputation. This dynamic exchange of information
allows the algorithm to regulate interaction intensity, ensuring both diversity preservation and efficient convergence.
Trust and reputation are terms that are being explored mostly in the fields of social psychology. Therefore, incorporating them into the algorithm makes it part of the socio-cognitive optimization framework \cite{byrskibook}.

The remainder of this paper is structured as follows: Section \ref{sec:tbo_orgin} discusses related work in evolutionary algorithms and multi-agent systems. Section \ref{sec:tbo_features} details the features of the trust-based optimization (TBO). Section \ref{sec:mathTBO} introduces the formal setup of the TBO algorithm, detailing its design and parameterization. Section~\ref{sec:experiments} presents experimental results that evaluate the effectiveness of our approach. Finally, Section \ref{sec:summary} concludes with future research directions.

\section{The origin and motivation for the TBO}\label{sec:tbo_orgin}

\enlargethispage{2.5\baselineskip}

An evolutionary algorithm is the first level upon which our proposed algorithm is built. It is a population‐based, stochastic optimization method inspired by natural evolution. Typically, EAs iteratively evolve multisets of candidate solutions using operators such as selection, crossover, and mutation, aiming to optimize a given fitness function \cite{holland1975adaptation}.

The island model is a parallel variant of evolutionary algorithms in which the overall population is partitioned into multiple, relatively isolated subpopulations (called islands) \cite{Schleuter1991Explicit,islandmodels}. These subpopulations evolve independently for a number of iterations and periodically exchange individuals through a migration process.
Although it is widely recognized that IM generally outperforms classical single-population EAs, particularly with appropriate parameter tuning~\cite{cantupaz}, there have been many attempts to improve it even further~\cite{Skolicki2004Improving,islandmodels,Lardeux2010Dynamic,Meng2017Dynamic}.


Multi-agent systems are frameworks for modeling decentralized processes, consisting of multiple autonomous entities (agents) that interact within a shared environment. Each agent operates based on its own objectives, and their interactions can be leveraged to solve complex problems.
An evolutionary multi-agent system is a hybrid framework that combines the principles of evolutionary algorithms and multi-agent systems. The way agents are defined within the system leads to two distinct architectures. In an evolutionary multi-agent system (EMAS), each agent represents a single individual \cite{emas}, whereas in a flock-based model (understood as in \cite{mkdFlock}), an agent oversees and manages a collective of individuals within an island. Hence, by adding a layer of multi-agent system over an island model, we end up with islands governed by independent agents that take control over some variation mechanisms. This type of algorithm architecture was also employed by Lopes et al. \cite{lopes}, who transform the IM into a multi-agent system,
generating and adapting the migration topologies with the use of a reinforcement learning method called Q-learning.

The Trust-Based Optimization (TBO) algorithm builds upon the evolutionary island model and multi-agent evolutionary systems by replacing the traditional migration phase with an adaptive interaction mechanism. Instead of periodic migration, agents exchange selected information based on learned trust levels or reputation scores, ensuring beneficial knowledge spreads while unreliable exchanges are filtered out.

This interaction mechanism allows for the integration of socio-cognitive elements and incorporates customized parameters, which will be described in detail in the following sections. 

\section{TBO features}\label{sec:tbo_features}

\subsection{Relations based on trust and reputation}\label{sec:trust-reputation}

The first component of inter-agent interaction is a dynamic relationship mechanism, which may be based either on \textbf{trust} or \textbf{reputation} (both variants are tested as part of different algorithm configurations). This mechanism aims to refine agents' interactions by rewarding reliable information sources and mitigating deceptive or low-quality exchanges.

Trust is a concept originating from the field of social psychology that refers to one's belief in the good intentions and positive attitudes of others. In practice, however, trust represents a wager or a bet taken on the uncertain future actions of other people, grounded in prior experiences and past interactions with them.


Within the algorithm, trust is modeled as a dynamic, pairwise relationship updated according to the outcome of interactions. After an exchange of information that enhances the average fitness of an agent's population, the agent increases its trust in the other agent by assigning an additional trust point. Conversely, if the received population's average fitness is significantly lower than the agent's current population fitness, the trust point is taken back.

On the other hand, reputation is an aggregate measure of an agent’s reliability as perceived by the entire system. Unlike trust---which represents a personalized, bilateral assessment---reputation is a publicly accessible score and reflects the  collective feedback received from multiple agents.

In the proposed algorithm, the reputation mechanism is implemented by allocating an initial number of tokens to each agent. Agents exchange these tokens throughout their interactions: an agent transfers tokens to another agent when the information received from that agent positively impacts its own performance. Consequently, the number of tokens an agent accumulates directly reflects its reputation within the system. By transferring tokens, agents collectively highlight reliable sources of information. Reputation scores are public, meaning that every agent has complete insight into the token counts of all other agents. As the algorithm proceeds, this mechanism is expected to encourage greater alignment with the global leader, promoting more intensive utilization of data obtained from the most reputable agents.

In summary, the level of trust or reputation between agents affects:
\begin{itemize}
\item the amount of information exchanged between agents,
\item the extend to which the received information is being utilized.
\end{itemize}

\subsection{Learning mechanism---socio-cognitive crossover}\label{sec:learning_mechanisms}
The second aspect of TBO involves the mechanism by which the information from one agent influences the population of the other. Based on the Social Learning Theory developed by Albert Bandura \cite{bandura1986}, people are learning through the observation of the actions of others and are drawing conclusions from their behavior. Incorporating this idea into the TBO, one agent is learning (shaping its own population) based on the observation made on the populations of the other agents. We propose a new learning operator that agents use in place of the conventional IM migration mechanism.
We named this operator a \textbf{socio-cognitive crossover} because
it updates existing genomes within an agent's population by incorporating information about other solutions received from the other agent.
Such a mechanism allows for control over the strength of the impact of the received population on the existing one.
The operator functions on two levels:
\begin{enumerate}
    \item \textbf{Genome level}
Defines the scope of modifications applied to individuals throughout the population.
\begin{itemize}
    \item \textbf{Weak} -- Incorporates single genes from received individuals into multiple genomes.
    \item \textbf{Moderate} -- Integrates all received individuals, assigning multiple genes across the population.
    \item \textbf{Strong} -- Extensively modifies multiple genomes by incorporating multiple genes from received individuals.
\end{itemize}
\item \textbf{Gene level}
Specifies the method of adjusting particular genes. The algorithm identifies genes exhibiting the greatest divergence and applies one of the two strategies:
\begin{itemize}
    \item \textbf{Swapping} -- Exchanges genes between individuals.
    \item \textbf{Averaging} -- Computes an average value for the genes and assigns it.
\end{itemize}
\end{enumerate}

\subsection{Interaction between agents}\label{sec:interactions}

\enlargethispage{2.5\baselineskip}

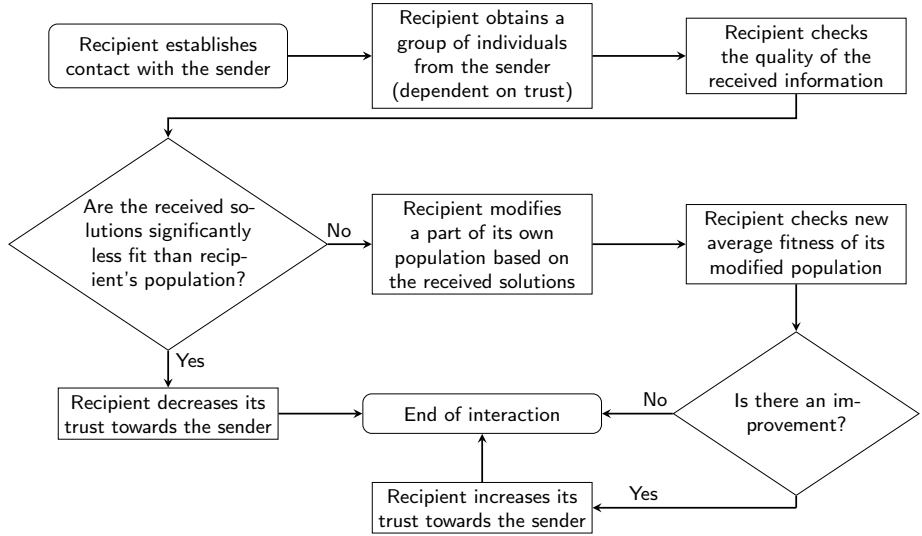
\begin{figure}[hbt!]
    \centering
    \makebox[\textwidth][c]{%
        \resizebox{1\textwidth}{!}{%
        \begin{tikzpicture}[>=stealth, auto, font=\sffamily]
    
            \pgfmathsetmacro{\xSpacingA}{5}
            \pgfmathsetmacro{\xSpacingB}{2 * \xSpacingA}
            \pgfmathsetmacro{\yLevelOne}{-3}
            \pgfmathsetmacro{\yLevelTwo}{1.9 * \yLevelOne}
            \pgfmathsetmacro{\shortOffset}{-1 * (\xSpacingA / 13)}
            \pgfmathsetmacro{\yLevelThree}{2.4 * \yLevelOne}
    
            \tikzstyle{startstop} = [
                rectangle,
                rounded corners,
                text width=3.4cm,
                align=center,
                draw=black,
                inner sep=2mm
            ]
            \tikzstyle{process} = [
                rectangle,
                text width=3.3cm,
                align=center,
                draw=black,
                inner sep=1mm
            ]
            \tikzstyle{decision} = [
                diamond,
                aspect=1.5,
                align=center,
                draw=black,
                text width=3.4cm,
                inner sep=-1mm
            ]
            \tikzstyle{arrow} = [thick,->,>=stealth]
    
            \node (start) [startstop] at (0,0)
                {Recipient establishes contact with the sender};
            \node (obtain) [process] at (\xSpacingA,0)
                {Recipient obtains a group of individuals from the sender (dependent on trust)};
            \node (check) [process] at (\xSpacingB,0)
                {Recipient checks the quality of the received information};
            \node (dec1) [decision] at (0,\yLevelOne)
                {Are the received solutions significantly less fit than recipient's population?};
            \node (end) [startstop] at (\xSpacingA,\yLevelTwo)
                {End of interaction};
            \node (adapt) [process] at (\xSpacingA,\yLevelOne)
                {Recipient modifies a part of its own population based on the received solutions};
            \node (select) [process] at (\xSpacingB,\yLevelOne)
                {Recipient checks new average fitness of its modified population};
            \node (dec2) [decision] at (\xSpacingB,\yLevelTwo)
                {Is there an improvement?};
            \node (trust) [process] at (\xSpacingA,\yLevelThree)
                {Recipient increases its trust towards the sender};
            \node (distrust) [process] at (0,\yLevelTwo)
                {Recipient decreases its trust towards the sender};
    
            \draw [arrow] (start) -- (obtain);
            \draw [arrow] (obtain) -- (check);
            \draw [arrow] (check.south)
                  -- ++(0,\shortOffset)
                  -- ++(-\xSpacingB,0)
                  -- (dec1.north);
            \draw [arrow] (dec1) -- node[near start] {No} (adapt);
            \draw [arrow] (dec1) -- node[near start] {Yes} (distrust);
            \draw [arrow] (adapt) -- (select);
            \draw [arrow] (select) -- (dec2);
            \draw [arrow] (distrust) -- (end);
            \draw [arrow] (dec2) -- node[near start, above] {No} (end);

            \coordinate (corner) at (dec2.south |- trust.east);
            \draw [-, thick] (dec2.south) -- (corner);
            \draw [arrow] (corner) -- (trust.east) node[near end, above] {Yes};

            \draw [arrow] (trust) --  (end);
    
        \end{tikzpicture}
        }
    }
    \caption{Schema of the interaction step illustrated using a trust-based relationship.}
    \label{fig:interaction}
\end{figure}

The core concept that distinguishes our algorithm from IMs is the interaction step (which we explain in greater detail
in the following section~\ref{sec:mathTBO} below). This step always involves two agents, where one agent requests information about the population of the other. The interaction occurs in the following steps, depicted also in Figure~\ref{fig:interaction}:
\begin{enumerate}
    \item At the beginning, the agent that will receive the information initiates the interaction by sending its ID to another agent and waiting for a response. The sender then verifies the recipient's credibility and provides data, adjusting the amount and quality based on the level of reputation/trust for the sender.
    \item Upon receiving the data, the recipient evaluates its quality by comparing the average fitness of the received individuals with that of its own population. If the received data falls significantly below an acceptable threshold, it is discarded, the subsequent interaction steps are skipped, and trust in the sender decreases.
    \item If the data passes verification, the recipient integrates a portion of the received population to modify its own. The degree of modification is determined by the recipient's trust in the sender or its reputation.
    \item After integrating the received information, the algorithm evaluates its impact on the recipient's population by comparing pre- and post-modification performance. If the interaction yields a beneficial outcome, the recipient increases its trust in the sender (or its reputation).
\end{enumerate}

\section{TBO algorithm}\label{sec:mathTBO}


Consider the TBO characterized by the tuple:
\begin{equation}
\Theta = \langle N, \mathcal{S}, f, \tau, d_{f}, R \rangle,
\label{eq:tbo}
\end{equation}
where $N$~is the size of a set of agents \( A = \{a_0, a_1, \dots, a_{N-1}\} \);
\( \mathcal{S} \subset \mathbb{R}^{\text{D}} \) is the search space of admissible solutions;
\( f: \mathcal{S} \to \mathbb{R} \) is a fitness function;
\( \tau = \{\tau_1, ..., \tau_n\} \): denotes the epoch durations (or interaction interval);
\(d_f\) is the diversity amplification factor;
and \( R \) is a configuration that specifies the type of the relations between agents (trust-based or reputation-based). The configuration also stores its initial and extreme values \( R = \langle \mathcal{C}_{\text{type}}, \mathcal{C}_{\text{start}}, \mathcal{C}_{\text{min}}, \mathcal{C}_{\text{max}}\rangle \).
Then each agent $a_i$
is a 6-tuple:
\begin{equation}
a_i = \langle n_i, \lambda_i, p_{c_i}, p_{m_i}, \mathcal{C}_i, X_{i} \rangle,
\label{eq:agent}
\end{equation}
where \(n_i\) is the size of the set of solutions called population \( \mathcal{P}_i \subset \mathcal{S} \), (i.e., \(|\mathcal{P}_i| = n_i\));
\(\lambda_i\) stands for offspring population size;
\(p_{c_i}\) and \(p_{m_i}\) represent crossover and mutation rates, respectively;
\(\mathcal{C}_i \in \big\{\{T_{j,i}\}, R_i \big\}\) is the social credibility component being either: a trust vector \(\{T_{j,i}\}\) with each \(T_{j,i} \in \mathbb{N}\) quantifying the trust that each agent \(a_j\) places in agent \(a_i\) (if using a trust-based configuration), or a reputation score \(R_i \in \mathbb{N}\) indicating the global trust of agent \(a_i\) (if using a reputation-based configuration);
\({X}_i\) denotes the configuration for socio-cognitive crossover operator \(\chi_{\text{SC}}\), with \({X}_i = \langle x_{\text{genome}}, x_{\text{gene}} \rangle\).

TBO framework resemblances IM. Each agent \(a_i\) independently manages a set of solutions (commonly referred to as individuals or genomes) that form the population \(P_i^{(t)} \subset \mathcal{S}\). This population evolves either through a standard evolutionary algorithm (EA) or via a specialized learning mechanism derived from socio-cognitive interactions between agents. The algorithm begins with the initialization of each agent's population and its social credibility component (trust levels vector or reputation level). Similar to the IM, during the interaction intervals (epochs), the solutions are evolved by the EA. At discrete time instants \(t = k\tau_i\) (with \(k \in \mathbb{N}\)), interactions occur that modify the populations through the socio-cognitive learning mechanism.

Formally, we define the global update rule as follows:
\begin{equation}
U: \prod_{i=1}^n \bigl(\mathcal{P}_i^{(t)} \times \mathcal{C}_i^{(t)}\bigr) \to \prod_{i=1}^n \bigl(\mathcal{P}_i^{(t+1)} \times \mathcal{C}_i^{(t+1)}\bigr),
\label{eq:update_rule}
\end{equation}
where \(\mathcal{P}_i^{(t)}\) denotes the population of agent \(a_i\) at time \(t\) and \(\mathcal{C}_i^{(t)}\) represents its associated social credibility component (trust or reputation). The state of each agent is updated according to:
\begin{equation}
\bigl(\mathcal{P}_i^{(t+1)},\, \mathcal{C}_i^{(t+1)},\, \mathcal{C}_j^{(t+1)}\bigr)
=
\begin{cases}
\text{EA}\bigl(\mathcal{P}_i^{(t)}\bigr) & \text{if } t \mod \tau_i \neq 0, \\
\text{I}\Bigl(\mathcal{P}_i^{(t)},\, \mathcal{P}_j^{(t)},\, \mathcal{C}_i^{(t)},\, \mathcal{C}_j^{(t)}\Bigr)
& \text{if } t \mod \tau_i = 0,
\end{cases}
\label{eq:update_agent}
\end{equation}
in which \(\text{EA}(\mathcal{P}_i^{(t)})\) denotes an evolutionary step that modifies \(\mathcal{P}_i^{(t)}\) through a combination of standard evolutionary operators, and \(\text{I}(\mathcal{P}_i^{(t)}, \mathcal{P}_j^{(t)}, \mathcal{C}_i^{(t)}, \mathcal{C}_j^{(t)})\) represents an interaction step based on a socio-cognitive learning mechanism. In the interaction step, an agent \(a_j\) is selected uniformly at random from the set \(A \setminus \{a_i\}\) to share its individuals with the agent \(a_i\).
Each interaction stage also updates the respective social credibility components (i.e., the trust or reputation levels) of the interacting agents. In reputation-based interactions, these values are updated for both agents; in trust-based interactions, only $\mathcal{C}_j^{(t)}$---more specifically \({T}_{ij}^{(t)}\), the trust that agent \(a_i\) assigns to agent \(a_j\)---is updated. When the EA step is executed (i.e., when \(t \mod \tau_i \neq 0\)), the social credibility components remain unchanged, so that \(\mathcal{C}_i^{(t+1)} = \mathcal{C}_i^{(t)}\) for any agent \(a_i \in A\).

\enlargethispage{2.5\baselineskip}

The evolutionary algorithm step \(\text{EA}(\mathcal{P}_i^{(t)})\) is conventionally defined as follows:
\begin{equation}
\text{EA}(\mathcal{P}_i^{(t)}) = \rho\Big(\mathcal{P}_i^{(t)}, \mathcal{O}_i^{(t)}\Big) \quad \text{with} \quad \mathcal{O}_i^{(t)} = \{ \mu(\chi(\sigma(\mathcal{P}_i^{(t)}, f))) \},
\label{eq:ae_step}
\end{equation}
where
$\rho$
is the replacement strategy,
\( \mu \)
is the mutation operator,
\( \chi \)
denotes the crossover operator,
\( \sigma \)
represents selection,
and $\mathcal{O}_i^{(t)}$ denotes the offspring population evolved from the population maintained by the agent $a_i$ at step $t$.
%
%
Additionally, in each population \(\mathcal{P}_i\), the crossover and mutation rates \(p_c\) and \(p_m\)
are further adjusted by the diversity amplification factor~\(d_f\):
\begin{equation}
p_c = p_{c_i} \bigl(1 + i\, d_f\bigr), \quad
p_m = p_{m_i} \bigl(1 + i\, d_f\bigr).
\label{eq:amplification_factor}
\end{equation}

The interaction step with the socio-cognitive learning mechanism is a two-phase process:
\begin{equation}
\text{I}\Bigl(
\mathcal{P}_i^{(t)},
\mathcal{P}_j^{(t)},
\mathcal{C}_{i}^{(t)},
\mathcal{C}_{j}^{(t)}
\Bigr)
=
\Bigl(
\text{L}\big(
\mathcal{P}_i^{(t)},
\mathcal{P}_j^{(t)},
\mathcal{C}_{i}^{(t)},
\mathcal{C}_{j}^{(t)}
\big)%
,
\upsilon\big(
\mathcal{P}_i^{(t)},
\mathcal{P}_j^{(t)},
\mathcal{C}_i^{(t)},
\mathcal{C}_j^{(t)}
\big)
\Bigr)
\label{eq:interaction_step}
\end{equation}
Here, L
denotes the learning phase, while $\upsilon$
updates the social credibility components.
Learning phase can specifically be defined as
\begin{equation}
\begin{aligned}
\mathcal{P}_i^{(t+1)} = \text{L}\Bigl(\mathcal{P}_i^{(t)},\,
\mathcal{P}_j^{(t)},\,
T_{j,i}^{(t)},\, T_{i,j}^{(t)}\Bigr) 
&= \rho\Bigl(\mathcal{P}_i^{(t)},\, \mathcal{O}_i^{(t)}\Bigr) \\
\quad\text{with}\quad
\mathcal{O}_i^{(t)} &= \mu_{SC}\Bigl(\mathcal{P}_i^{(t)},\,
\sigma_{SC}\bigl(\mathcal{P}_j^{(t)}, f, T_{j,i}^{(t)}\bigr),\,
T_{i,j}^{(t)}\Bigr)
\end{aligned}
\label{eq:learining_phase}
\end{equation}
for the trust-based configuration, and 
similarly for reputation-based configuration,\footnote{In most subsequent definitions, the choice between a trust-based or reputation-based configuration does not significantly affect the formulation. For convenience, we adopt trust-based notation, which can be readily adjusted for reputation-based configurations by replacing the trust term (e.g., \(T_{ji}\)) with the reputation term (e.g., \(R_i\)). The resulting formulae would remain equivalent \textit{mutatis mutandis}.}
where
\(\rho\) denotes the replacement strategy, while \(\mu_{SC}\) and \(\sigma_{SC}\) represent the socio-cognitive variation and selection operators, respectively. Both operators rely heavily on the social credibility component (i.e., trust or reputation).
It is convenient to refer to \(a_i\) as the \textbf{recipient} and \(a_j\) as the \textbf{sender} during the interaction process.

\enlargethispage{2.5\baselineskip}

At the first stage of the learning phase, a selection operator \( \sigma_{SC} : \mathcal{S} \times \mathbb{R} \times \mathbb{N} \to \mathcal{S} \) is employed to select a subset of \(\mathcal{P}_j^{(t)}\) based on the fitness of its individuals and the social credibility component $\mathcal{C}_i$. The trust level that the sender assigns to the recipient (or, in another configuration, the recipient's reputation) determines the number of candidate solutions shared with $a_i$. Specifically, a solution $y$ is included in the shared population $\mathcal{Q}_{ji}^{(t)} \subset \mathcal{P}_j^{(t)}$ if and only if $y$ is among the $T_{j,i}^{(t)}$ (or \(R_{i}^{(t)}\)) least fit individuals---that is, those solutions with the highest $f(y)$ values (assuming minimization). Formally, we define
\textls[-5]{\begin{equation}
\!\mathcal{Q}_{ji}^{(t)}\!=\!
\sigma_{SC}\Big(\mathcal{P}_j^{(t)}, f, T_{j,i}^{(t)}\Big)\!=\!
\left\{ y \in \mathcal{P}_j^{(t)} \,\bigg|\left|\left\{ z \in \mathcal{P}_j^{(t)}\!\mid f(z)\!<\!f(y) \right\}\right|\!\geq\! N_j\!-\!T_{j,i}^{(t)} \right\}
\label{eq:shared_population}
\end{equation}}
Note that with this formulation, a higher value of $T_{ji}$ results in the selection of higher-quality solutions from $\mathcal{P}_j^{(t)}$ for sharing.

\textls[-10]{Next, the socio-cognitive variation operator \(\mu_{SC}\) is applied. The operator uses the recipient’s population \(\mathcal{P}_i^{(t)}\), the shared population \(\mathcal{Q}_{ji}^{(t)}\), and the trust level \(T_{i,j}^{(t)}\) that the recipient assigns to the sender to generate the offspring population \(\mathcal{O}_i^{(t)}\):}
\begin{equation}
\mathcal{O}_i^{(t)} \!=\!
\mu_{SC}(\mathcal{P}_i^{(t)},
\mathcal{Q}_{ji}^{(t)},
T_{i,j}^{(t)}) 
\!=\!
\begin{cases}
\mathcal{P}_i^{(t)},\!&\!\text{if } \bar{f}\big(\mathcal{Q}_{ji}^{(t)}\big) > \epsilon_i(\mathcal{P}_i^{(t)}), \\
\chi_{SC}(\mathcal{P}_i^{(t)},
\mathcal{Q}_{ji}^{(t)},
T_{i,j}^{(t)})\!&\!\text{otherwise},
\end{cases}
\label{eq:offspring_population}
\end{equation}
where \(\chi_{SC}\) denotes the socio-cognitive crossover operator, and \(\bar{f}\)
denotes the average fitness of a given population. Various acceptance thresholds can be employed. One possible definition is as follows:
\begin{equation}
\epsilon_i(\mathcal{P}_i^{(t)}) =
\begin{cases}
2\,\bar{f}\big(\mathcal{P}_i^{(t)}\big) & \text{if } \bar{f}\big(\mathcal{P}_i^{(t)}\big) > 0, \\
0 & \text{otherwise}.
\end{cases}
\label{eq:acceptance_treshold}
\end{equation}

We assume each candidate solution (or genome) is represented as
\begin{equation}
y = \bigl(y_1, y_2, \dots, y_D\bigr) \in \mathcal{S} \subset \mathbb{R}^D,
\label{eq:genome}
\end{equation}
Socio-cognitive crossover operator $\chi_{SC}: \mathcal{S} \times \mathcal{S} \times \mathbb{N} \times X_i \to \mathcal{S}$
acts on a solution \(y \in \mathcal{Q}_{ji}^{(t)}\) using information from a solution \(x \in \mathcal{P}_i^{(t)}\) and a trust level \(T_{i,j}^{(t)}\). The operator configuration \(X_i = \langle x_{\text{genome}}, x_{\text{gene}} \rangle\) determines both the mutation intensity (weak, moderate or strong) and the gene‐level operation (swap or average).
The operator may be defined by altering the genes in \( y \) based on the most divergent genes in \( x \).

For each gene  \(y_i, i \in \{1,\dots,D\}\), define the absolute difference
\(
d_i = \bigl|x_i - y_i\bigr|
\).
Assume indices are sorted in descending order according to \(d_i\) , and let
\(
\mathcal{I}(k) \subseteq \{1,2,\dots,D\}
\)
denote the set of indices corresponding to the \(k\) largest differences.
Then, the socio‐cognitive crossover operator is defined by modifying the genes in \(y\) according to  
\begin{equation}
\phi(y_i, x_i, k) =
\begin{cases}
x_i, & \text{if } i \in \mathcal{I}(k) \text{ and } x_{\text{gene}} = \texttt{swap}, \\
\displaystyle \frac{y_i + x_i}{2}, & \text{if } i \in \mathcal{I}(k) \text{ and } x_{\text{gene}} = \texttt{average}, \\
y_i, & \text{otherwise}.
\end{cases}
\label{eq:sccrossover}
\end{equation}

Let
\(
K = \min\{T_{i,j}^{(t)},\, D\}.
\)
Based on this, we define three mutation schemes:
\begin{itemize}
    \item Weak:
   \( \forall\, y \in \mathcal{Q}_{ji}^{(t)},\; \exists\, x \in \mathcal{P}_i^{(t)}:\; y' = \phi(y, x, K).\)
    \item Moderate:
    \( \forall\, y \in \mathcal{Q}_{ji}^{(t)},\; \forall\, k \in \{1,\dots,K\}:\; y^{(k)} = \phi(y, x, K).\)
    \item Strong:
   \( \forall\, y \in \mathcal{Q}_{ji}^{(t)},\; \forall\, k \in \{1,\dots,K\}:\; y^{(k)} = \phi(y, x, 1).\)
\end{itemize}

\enlargethispage{2.5\baselineskip}

In the final phase of the interaction, the social credibility component is updated according to
\begin{equation}
(\mathcal{C}_i^{(t+1)},
\mathcal{C}_j^{(t+1)}) =
\upsilon\big(
\mathcal{P}_i^{(t)},
\mathcal{P}_j^{(t)},
\mathcal{C}_i^{(t)},
\mathcal{C}_j^{(t)}
\big)
\label{eq:credibility_update}
\end{equation}
Specifically, 
the trust $T_{i,j}$ is updated as:
\begin{equation}
T_{i,j}^{(t+1)} =
\begin{cases}
T_{i,j}^{(t)} + 1, & \text{if } \bar{f}\big(\mathcal{P}_i^{(t+1)}\big) < \bar{f}\big(\mathcal{P}_i^{(t)}\big) 
\\
\max\{1,\, T_{i,j}^{(t)} - 1\}, & \text{if } \bar{f}\big(\mathcal{Q}_{ji}^{(t)}\big) > \epsilon_i(t) 
\\
T_{i,j}^{(t)}, & \text{otherwise}, 
\end{cases}
\label{eq:trust_update}
\end{equation}
and no other social credibility component is being updated during trust-based interactions. For the reputation-based configuration, both agents' reputation scores are updated as:
\begin{equation}
\begin{array}{c}
(R_i^{(t+1)}, R_j^{(t+1)}) = \\
\begin{cases}
(\max\{1,\, R_i^{(t)} - 1\}, \min\{R_{\text{max}},\, R_j^{(t)} + 1\}), & \text{if } \bar{f}\bigl(\mathcal{P}_i^{(t+1)}\bigr) < \bar{f}\bigl(\mathcal{P}_i^{(t)}\bigr), \\
(\min\{R_{\text{max}},\, R_i^{(t)} + 1\}, \max\{1,\, R_j^{(t)} - 1\}), & \text{if } \bar{f}\bigl(\mathcal{Q}_{ji}^{(t)}\bigr) > \epsilon_i\bigl(\mathcal{P}_i^{(t)}\bigr), \\
(R_i^{(t+1)}, R_j^{(t+1)}), & \text{otherwise}.
\end{cases}
\end{array}
\label{eq:reputation_update}
\end{equation}
where \(R_{\text{max}}\) denotes the maximum reputation score.

The general overview of the algorithm is presented in the schema of Algorithm~\ref{alg:TBO}.

\begin{algorithm}[htbp]
\SetKwInOut{Input}{Input}\SetKwInOut{Output}{Output}
\Input{\(N, \mathcal{S}, f, \tau, d_{f}, R\). For each agent
\(a_i \in A\),
\( n_i, \lambda_i, p_{c_i}, p_{m_i}, \mathcal{C}_i, X_{i}\).}
\Output{Best solution found}
\textbf{Initialize:} For each agent \(a_i\), set population \(\mathcal{P}_i^{(0)}\) and credibility \(\mathcal{C}_i^{(0)}\). Adjust \(p_{c_i}\) and \(p_{m_i}\) with \(f_d\). Set \(t \gets 0\).\;
\While{Termination criterion is not met}{
  \For{each agent \(a_i \in A\)}{
    \If{\(t \mod \tau_i \neq 0\)}{
      Perform evolutionary step \(\text{EA}(\mathcal{P}_i^{(t)})\).\;
    }
    \Else{
      \tikz[remember picture,overlay] \node (ElseStart) {};%
      Randomly select a sender \(a_j \in A \setminus \{a_i\}\).\;
      Select individuals for shared population \(\mathcal{Q}_{ji}^{(t)} \gets \sigma_{SC}\Big(\mathcal{P}_j^{(t)}, f, \mathcal{C}_{i}^{(t)}\Big) \).\;
      Perform socio-cognitive variation operation to generate offspring population \(\mathcal{O}_i^{(t)} \gets \mu_{SC}(\mathcal{P}_i^{(t)},\mathcal{Q}_{ji}^{(t)},\mathcal{C}_j^{(t)}) \).\;
      Replace the population \(\mathcal{P}_i^{(t+1)} \gets \rho(\mathcal{P}_i^{(t)}, \mathcal{O}_i^{(t)})\).\;
      \tikz[remember picture,overlay] \node (ElseEnd) {};\tikz[remember picture,overlay]{
        \draw[decorate,decoration={brace,amplitude=10pt,raise=4pt}]
          ([xshift=93mm]ElseStart.north east) -- ([xshift=93mm]ElseEnd.south east)
          node[midway,xshift=7.7mm,rotate=90,scale=0.8]{Interaction step};
      }%
      Update \(\mathcal{C}_i\) and \(\mathcal{C}_j\), if necessary.\;
    }%
  }%
  \(t \gets t + 1\).\;
}
\Return The optimal solution across all populations.\;
\caption{Trust-Based Optimization (TBO) Algorithm}
\label{alg:TBO}
\end{algorithm}

\section{Experiments and results}\label{sec:experiments}

\enlargethispage{2.5\baselineskip}

\subsection{Baseline algorithm and parameters values}
We adopt the IM as the baseline algorithm, as it serves as a foundational framework for TBO. Since both approaches consist of multiple independent EA populations, we first outline the parameter settings specific to the evolutionary algorithm, followed by those related to the island configuration, and finally, the TBO-specific parameter values.
\\
\textbf{EA parameters}
-- variation operators and parameters used for the evolutionary algorithm: 
population size
\(n = 5\);
offspring size
\(\lambda = 15\);
selection
\(\sigma \): binary tournament selection;
crossover
\(\chi\): simulated binary crossover;
mutation type
\(\mu\):~polynomial mutation with distribution index $\eta_m = 40$;
crossover rate
\(p_c\): varies across islands, starting from 0.005 and increasing by a diversity factor $d_f$;
 mutation rate
 \(p_m\): varies across islands, starting from 0.0005 and increasing by a diversity factor $d_f$;
 replacement strategy
\( \rho\): ($\mu + \lambda$) with elitism.
\\
\textbf{IM parameters}
-- parameters of operators added on the island model level:
number of islands/agents
\(N \in \{ 5, 10, 20\}\);
epoch duration
\(\tau \in \{ 25, 50 \}\);
migration rate: 1  (baseline) or depending on reputation/trust (TBO),
migrant selection: elitist (baseline) or depending on reputation/trust (TBO).
diversity amplification factor
$d_f \in \{1.3, 2\}$.
\\
\textbf{TBO parameters}
-- summary of the possible values of socio-cognitive parameters, described in detail in Sections~\ref{sec:tbo_features} and~\ref{sec:mathTBO}:
relation
$\mathcal{C}_{\text{type}} \in \{ \text{reputation}, \text{trust} \}$,
$\mathcal{C}_{\text{start}} \in \{5, 25, 30, 40, 50 \}$;
$\mathcal{C}_{\text{min}} = 1$;
\mbox{S-C} crossover intensity
$x_{\text{genome}}  \in \{ \text{weak}, \allowbreak \text{moderate}, \allowbreak \text{strong} \}$
gene change
$x_{\text{gene}} \in \{ \text{swap}, \allowbreak \text{average} \}$.


Five distinct TBO-specific parameter configurations were selected for experimental evaluation. Each configuration highlights particular characteristics of the algorithm, leading to varying performance across different problem types, in accordance with the No Free Lunch Theorem \cite{noFreeLunch}. The parameter values for these configurations are collected in Table~\ref{tab:configs}. In our experimental setup, all agents within a single TBO system were configured with the same parameter values. However, in general, this homogeneity is not a requirement. In particular, interaction steps do not need to be synchronous.

\begin{description}
    \item[Strong leadership] -- determines the effectiveness of interactions based on high initial reputation along with intensive exchange of information.
    \item[Exploration] -- determines the effectiveness of interactions based on extensive exploration of the solution space by using  trust together with the average gene-change operator.
    \item[Small society] -- determines the effectiveness of interactions based on trust with a smaller number of agents.
    \item[Large society] -- determines the effectiveness of interactions based on reputation with a greater number of agents.
    \item[High diversity] -- determines the effectiveness of interactions based on intensive exchange of information with high diversity in the agent population. 
\end{description}

\begin{table}[htbp]
\caption{Set of configurations used during experiments.}
\label{tab:configs}
\centering
\begin{tabular}{ |c|c|c|c|c|c| }
\hline
\makecell{\textbf{Parameter}} & \makecell{\textbf{Strong}\\\textbf{leadership}} & \makecell{\textbf{Explo-}\\\textbf{-ration}} & \makecell{\textbf{Small}\\\textbf{society}} & \makecell{\textbf{Large}\\\textbf{society}} & \makecell{\textbf{High}\\\textbf{diversity}}
\\
\hline
$N$ & 10 & 10 & 5 & 20 & 10\\
\hline
$\tau$ & 25 & 25 & 25 & 50 & 25\\
\hline
$\mathcal{C}_{\text{type}}$ & reputation & trust & trust & reputation & reputation\\
\hline
$\mathcal{C}_{\text{start}}$ & 50 & 25& 5 & 30 & 40\\
\hline
$x_{\text{genome}}$ & moderate & strong & strong & weak & moderate\\
\hline
$x_{\text{gene}}$ & swap & average & swap & swap & swap\\
\hline
$d_f$ & 1.3 & 1.3& 1.3 & 1.3 & 2\\
\hline

\end{tabular}
\end{table}

These configurations were compared against a reference island model evolutionary algorithm to assess improvements in performance.


\enlargethispage{2.5\baselineskip}

\subsection{Experiment setup}\label{sec:experiments_setup}
The experiments were conducted to evaluate the performance of the proposed TBO algorithm in solving various optimization problems. The progress of the algorithm is assessed in terms of minimizing the mean value of the objective function as the number of iterations performed by the evolutionary algorithm progresses. The algorithms have been implemented using jMetalPy computing framework \cite{jMetalPy}.
 
The experiments were conducted on six optimization problems, listed below. 
\begin{flalign}
& f(x) = \sum_{i=1}^{d} x_i^2 && \tag{Sphere} \displaybreak[1] \\
& f(x) = \frac{1}{4000} \sum_{i=1}^{d} x_i^2 - \prod_{i=1}^{d} \cos\left(\frac{x_i}{\sqrt{i}}\right) + 1 && \tag{Griewank} \displaybreak[1] \\
& f(x) = \sum_{i=1}^{d} \Bigl[ x_i^2 - 10\cos(2\pi x_i) + 10 \Bigr] && \tag{Rastrigin} \displaybreak[1] \\
& f(x) = \sum_{i=1}^{d-1} \left[ 0.5 + \frac{\sin^2\Bigl(\sqrt{x_i^2+x_{i+1}^2}\Bigr)-0.5}{\left(1+0.001\,(x_i^2+x_{i+1}^2)\right)^2} \right] && \tag{Expanded Schaffer} \displaybreak[1] \\
& f(x) = 418.9829d - \sum_{i=1}^{d} x_i \sin\bigl(\sqrt{|x_i|}\bigr) + \epsilon, && \tag{Schwefel with Noise} \displaybreak[1] \\
&\text{where }  \epsilon  \text{ is a small random noise term.}  &&\nonumber \displaybreak[1] \\
&f(r) = \sum_{i=1}^{d-1} \sum_{j=i+1}^{d} \left[ \frac{A}{r_{ij}^{12}} - \frac{B}{r_{ij}^{6}} \right], && \tag{Lennard-Jones Minimum Energy Cluster} \displaybreak[1] \\
& \text{where \( r_{ij} \) is the distance between two particles, and \( A, B \) are constants defining} && \nonumber \\
& \text{the depth and equilibrium distance of the potential well.} && \nonumber
\end{flalign}
Most problems were tested with three different problem sizes: 50, 100, and 200 dimensions. The number of algorithm iterations was set according to the problem size, with 100 000 iterations for 50 dimensions, 200 000 iterations for 100 dimensions, and 400 000 iterations for 200 dimensions. The Lennard-Jones Minimum Energy Cluster problem was tested only for problem sizes of 50 and 100 dimensions. The experiments were repeated eight times on each of the problem dimensions. 

\subsection{Experimental results}\label{sec:experiment_summary}

\enlargethispage{2.5\baselineskip}

\begin{figure}[hbt!]
    \centering
    \begin{subfigure}[b]{0.49\textwidth}
        \centering
        \includegraphics[width=1\textwidth]{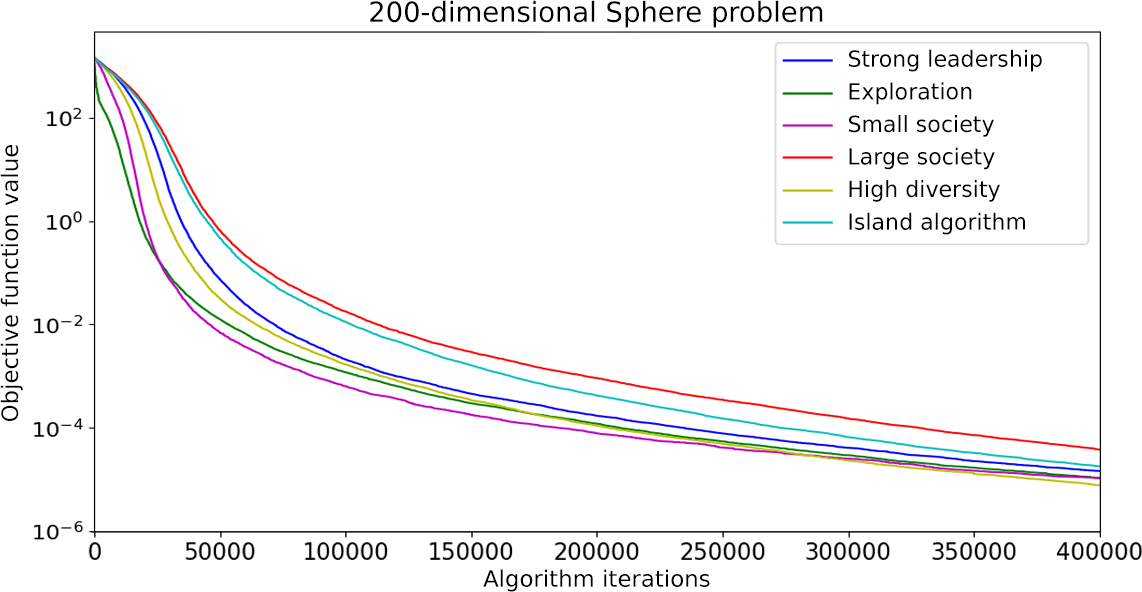}
        \caption{}
        \label{fig:sphere}
    \end{subfigure}
    \begin{subfigure}[b]{0.49\textwidth}
        \centering
        \includegraphics[width=1\textwidth]{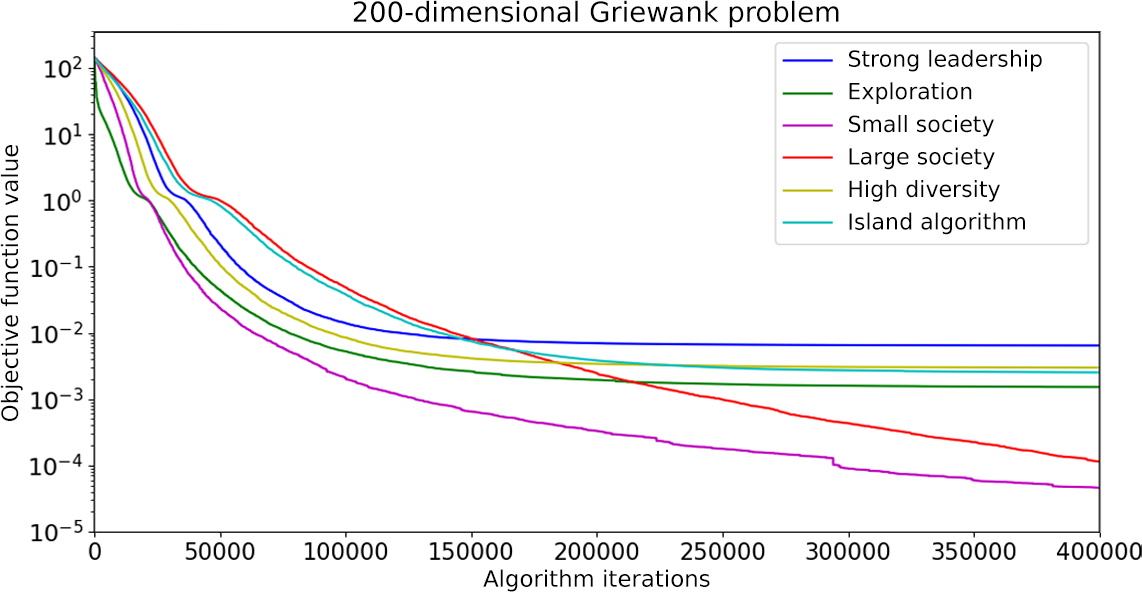}
        \caption{}
        \label{fig:griewank}
    \end{subfigure}
    \begin{subfigure}[b]{0.49\textwidth}
        \centering
        \includegraphics[width=1\textwidth]{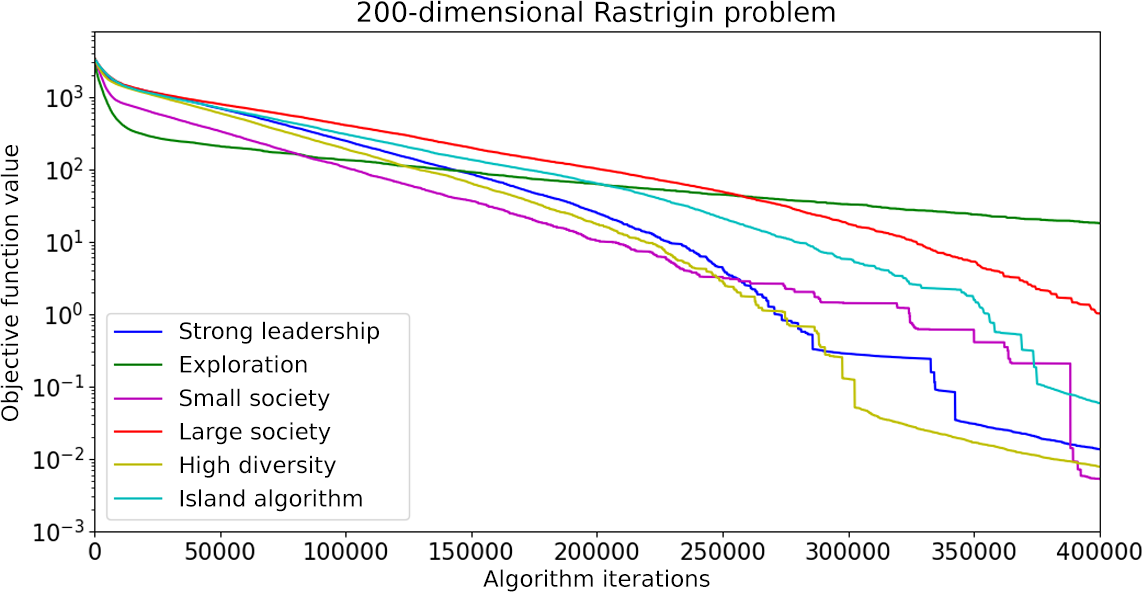}
        \caption{}
        \label{fig:rastrigin}
    \end{subfigure}
    \begin{subfigure}[b]{0.49\textwidth}
        \centering
        \includegraphics[width=1\textwidth]{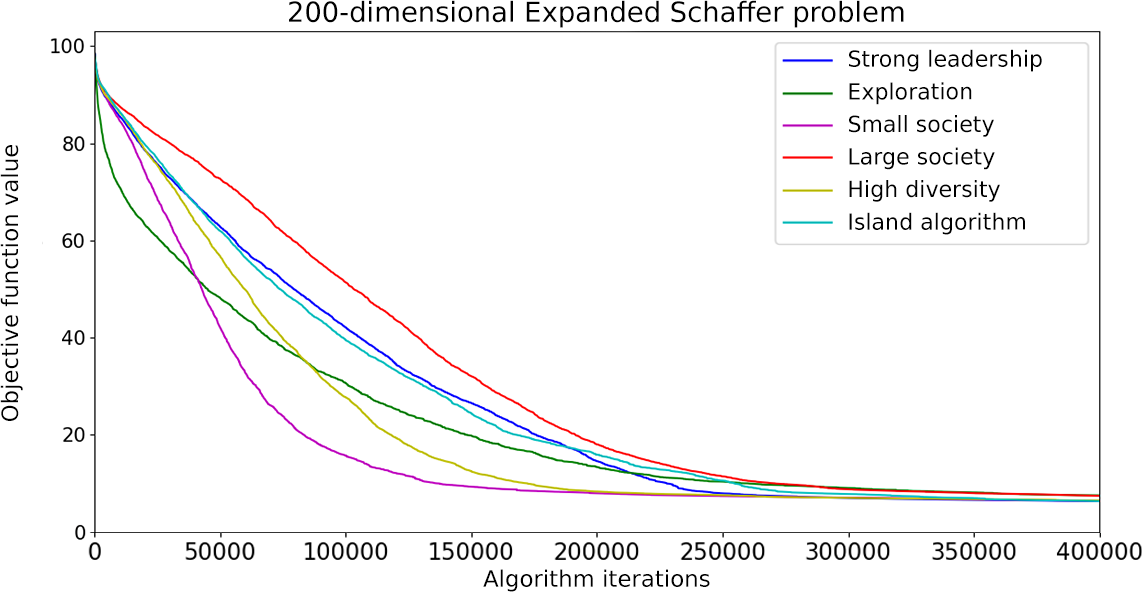}
        \caption{}
        \label{fig:shaffer}
    \end{subfigure}
    \begin{subfigure}[b]{0.49\textwidth}
        \centering
        \includegraphics[width=1\textwidth]{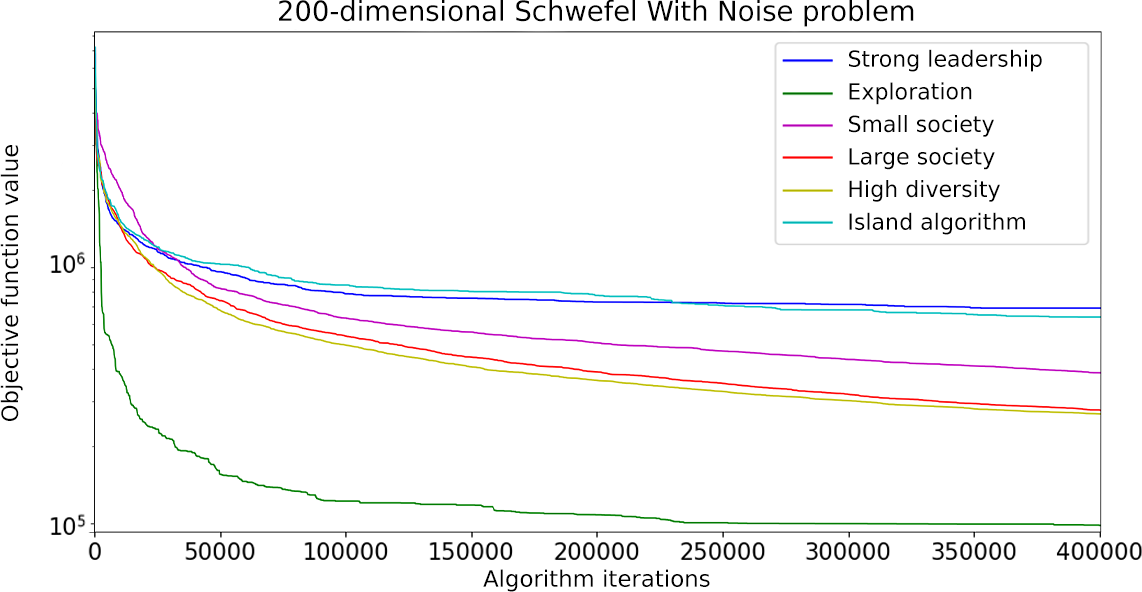}
        \caption{}
        \label{fig:schwefel}
    \end{subfigure}
    \begin{subfigure}[b]{0.49\textwidth}
        \centering
        \includegraphics[width=1\textwidth]{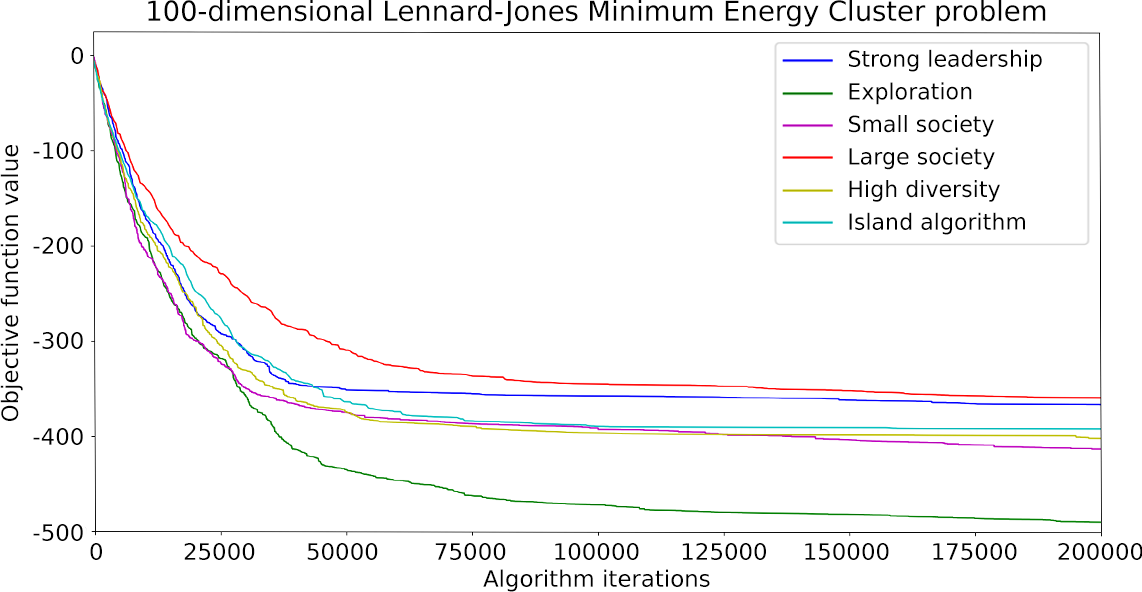}
        \caption{}
        \label{fig:ljmec}
    \end{subfigure}
    \caption{Convergence analysis of 
            \textit{Strong leadership} (\textcolor{blue}{blue}), 
            \textit{Exploration} (\textcolor{OliveGreen}{green}), 
            \textit{Small society} (\textcolor{RedViolet}{magenta}), 
            \textit{Large society} (\textcolor{red}{red}),
            and
            \textit{High diversity} (\textcolor{Goldenrod}{yellow})
            TBO configurations,
            as well as Island model (\textcolor{cyan}{cyan}) 
            on six benchmark functions---200-di\-men\-sional Sphere (a), Griewank (b), 
            Rastrigin (c), Expanded Shaffer (d), 
            Schwefel with noise problem (e) evaluated over 400,000 iterations 
            and 100-dimensional Lennard-Jones Minimum Energy Cluster problem (f) 
            evaluated over 200,000 iterations.}
    \label{fig:highest_dims_plots}
\end{figure}

The comparative analysis of different algorithm configurations revealed that the socio-cognitive mutation operator improved optimization performance across each tested benchmark problems. The mean fitness values and standard deviations are listed in Table \ref{tab:all_results_table}, followed by convergence graphs for the highest dimensionalities of the benchmark problems in Figure \ref{fig:highest_dims_plots}. The results indicate that the trust-based learning-inspired approach enhances convergence.


\begin{table}[ht!]
\centering
\caption{The experimental results for all TBO configurations and IM across all tested benchmark problems. Each cell contains the mean fitness value on the top and the standard deviation (SD) on the bottom. Bold font highlights the algorithm that achieved the best result.}
\label{tab:all_results_table}
\resizebox{\textwidth}{!}{%
\begin{tabular}{|c|c|c|c|c|c|c|c|c|}
\hline
\makecell{\textbf{Problem}} & \makecell{\textbf{Dim.}} & \makecell{\textbf{Iter.}} & \makecell{\textbf{Strong}\\\textbf{leadership}} & \makecell{\textbf{Explo-}\\\textbf{-ration}} & \makecell{\textbf{Small}\\\textbf{society}} & \makecell{\textbf{Large}\\\textbf{society}} & \makecell{\textbf{High}\\\textbf{diversity}} & \makecell{\textbf{Island}\\\textbf{model}} \\
\hline
\multirow{5}{*}{\centering\shortstack{\textbf{Sphere}}} 
 & \makecell{\shortstack{50\\\ }}  & \makecell{\shortstack{100000\\\ }} 
   & \shortstack{$2.0\cdot10^{-5}$\\{\scriptsize$\pm1.4\cdot10^{-6}$}} 
   & \shortstack{$1.5\cdot10^{-5}$\\{\scriptsize$\pm1.2\cdot10^{-6}$}} 
   & \shortstack{$2.9\cdot10^{-5}$\\{\scriptsize$\pm8.0\cdot10^{-7}$}} 
   & \shortstack{$2.4\cdot10^{-5}$\\{\scriptsize$\pm1.5\cdot10^{-6}$}} 
   & \shortstack{$\mathbf{8.1\cdot10^{-6}}$\\{\scriptsize$\pm5.5\cdot10^{-7}$}} 
   & \shortstack{$2.9\cdot10^{-4}$\\{\scriptsize$\pm1.1\cdot10^{-5}$}} \\
\cline{2-9}
 & \makecell{\shortstack{100\\\ }} & \makecell{\shortstack{200000\\\ }} 
   & \shortstack{$1.3\cdot10^{-5}$\\{\scriptsize$\pm6.1\cdot10^{-7}$}} 
   & \shortstack{$1.1\cdot10^{-5}$\\{\scriptsize$\pm4.8\cdot10^{-7}$}} 
   & \shortstack{$1.5\cdot10^{-5}$\\{\scriptsize$\pm5.9\cdot10^{-7}$}} 
   & \shortstack{$2.3\cdot10^{-5}$\\{\scriptsize$\pm6.2\cdot10^{-7}$}} 
   & \shortstack{$\mathbf{5.6\cdot10^{-6}}$\\{\scriptsize$\pm4.1\cdot10^{-7}$}} 
   & \shortstack{$1.2\cdot10^{-5}$\\{\scriptsize$\pm7.8\cdot10^{-7}$}} \\
\cline{2-9}
 & \makecell{\shortstack{200\\\ }} & \makecell{\shortstack{400000\\\ }} 
   & \shortstack{$1.5\cdot10^{-5}$\\{\scriptsize$\pm6.6\cdot10^{-7}$}} 
   & \shortstack{$1.1\cdot10^{-5}$\\{\scriptsize$\pm4.5\cdot10^{-7}$}} 
   & \shortstack{$1.1\cdot10^{-5}$\\{\scriptsize$\pm2.8\cdot10^{-7}$}} 
   & \shortstack{$3.8\cdot10^{-5}$\\{\scriptsize$\pm7.4\cdot10^{-7}$}} 
   & \shortstack{$\mathbf{7.7\cdot10^{-6}}$\\{\scriptsize$\pm2.1\cdot10^{-7}$}} 
   & \shortstack{$1.8\cdot10^{-5}$\\{\scriptsize$\pm1.9\cdot10^{-7}$}} \\
\hline
\multirow{5}{*}{\centering\shortstack{\textbf{Griewank}}} 
 & \makecell{\shortstack{50\\\ }}  & \makecell{\shortstack{100000\\\ }} 
   & \shortstack{$1.5\cdot10^{-2}$\\{\scriptsize$\pm2.6\cdot10^{-3}$}} 
   & \shortstack{$5.7\cdot10^{-3}$\\{\scriptsize$\pm2.3\cdot10^{-3}$}} 
   & \shortstack{$2.9\cdot10^{-2}$\\{\scriptsize$\pm1.4\cdot10^{-2}$}} 
   & \shortstack{$\mathbf{2.4\cdot10^{-3}}$\\{\scriptsize$\pm1.2\cdot10^{-3}$}} 
   & \shortstack{$2.1\cdot10^{-2}$\\{\scriptsize$\pm1.2\cdot10^{-2}$}} 
   & \shortstack{$8.0\cdot10^{-3}$\\{\scriptsize$\pm2.2\cdot10^{-3}$}} \\
\cline{2-9}
 & \makecell{\shortstack{100\\\ }} & \makecell{\shortstack{200000\\\ }} 
   & \shortstack{$1.3\cdot10^{-2}$\\{\scriptsize$\pm6.3\cdot10^{-3}$}} 
   & \shortstack{$7.3\cdot10^{-5}$\\{\scriptsize$\pm3.5\cdot10^{-6}$}} 
   & \shortstack{$1.2\cdot10^{-2}$\\{\scriptsize$\pm5.7\cdot10^{-3}$}} 
   & \shortstack{$7.0\cdot10^{-3}$\\{\scriptsize$\pm2.5\cdot10^{-3}$}} 
   & \shortstack{$\mathbf{5.3\cdot10^{-5}}$\\{\scriptsize$\pm3.7\cdot10^{-6}$}} 
   & \shortstack{$1.6\cdot10^{-3}$\\{\scriptsize$\pm8.9\cdot10^{-4}$}} \\
\cline{2-9}
 & \makecell{\shortstack{200\\\ }} & \makecell{\shortstack{400000\\\ }} 
   & \shortstack{$6.5\cdot10^{-3}$\\{\scriptsize$\pm2.9\cdot10^{-3}$}} 
   & \shortstack{$1.5\cdot10^{-3}$\\{\scriptsize$\pm8.9\cdot10^{-4}$}} 
   & \shortstack{$\mathbf{4.6\cdot10^{-5}}$\\{\scriptsize$\pm4.6\cdot10^{-6}$}} 
   & \shortstack{$1.1\cdot10^{-4}$\\{\scriptsize$\pm3.8\cdot10^{-6}$}} 
   & \shortstack{$3.0\cdot10^{-3}$\\{\scriptsize$\pm1.8\cdot10^{-3}$}} 
   & \shortstack{$2.5\cdot10^{-3}$\\{\scriptsize$\pm1.5\cdot10^{-3}$}} \\
\hline
\multirow{5}{*}{\centering\shortstack{\textbf{Rastrigin}}} 
 & \makecell{\shortstack{50\\\ }}  & \makecell{\shortstack{100000\\\ }} 
   & \shortstack{$2.1\cdot10^{0}$\\{\scriptsize$\pm4.1\cdot10^{-1}$}} 
   & \shortstack{$1.4\cdot10^{1}$\\{\scriptsize$\pm6.8\cdot10^{-1}$}} 
   & \shortstack{$4.6\cdot10^{0}$\\{\scriptsize$\pm6.5\cdot10^{-1}$}} 
   & \shortstack{$1.2\cdot10^{0}$\\{\scriptsize$\pm1.9\cdot10^{-1}$}} 
   & \shortstack{$\mathbf{4.1\cdot10^{-1}}$\\{\scriptsize$\pm1.5\cdot10^{-1}$}} 
   & \shortstack{$4.3\cdot10^{-1}$\\{\scriptsize$\pm2.4\cdot10^{-1}$}} \\
\cline{2-9}
 & \makecell{\shortstack{100\\\ }} & \makecell{\shortstack{200000\\\ }} 
   & \shortstack{$4.2\cdot10^{-1}$\\{\scriptsize$\pm1.5\cdot10^{-1}$}} 
   & \shortstack{$1.5\cdot10^{1}$\\{\scriptsize$\pm4.4\cdot10^{-1}$}} 
   & \shortstack{$8.2\cdot10^{-1}$\\{\scriptsize$\pm2.3\cdot10^{-1}$}} 
   & \shortstack{$4.9\cdot10^{-1}$\\{\scriptsize$\pm1.4\cdot10^{-1}$}} 
   & \shortstack{$\mathbf{3.0\cdot10^{-2}}$\\{\scriptsize$\pm1.4\cdot10^{-2}$}} 
   & \shortstack{$2.4\cdot10^{-1}$\\{\scriptsize$\pm1.2\cdot10^{-1}$}} \\
\cline{2-9}
 & \makecell{\shortstack{200\\\ }} & \makecell{\shortstack{400000\\\ }} 
   & \shortstack{$1.4\cdot10^{-2}$\\{\scriptsize$\pm5.3\cdot10^{-4}$}} 
   & \shortstack{$1.8\cdot10^{1}$\\{\scriptsize$\pm8.8\cdot10^{-1}$}} 
   & \shortstack{$\mathbf{5.3\cdot10^{-3}}$\\{\scriptsize$\pm3.5\cdot10^{-4}$}} 
   & \shortstack{$1.0\cdot10^{0}$\\{\scriptsize$\pm2.3\cdot10^{-1}$}} 
   & \shortstack{$7.8\cdot10^{-3}$\\{\scriptsize$\pm4.6\cdot10^{-4}$}} 
   & \shortstack{$5.9\cdot10^{-2}$\\{\scriptsize$\pm5.3\cdot10^{-3}$}} \\
\hline
\multirow{5}{*}{\centering\shortstack{\textbf{Expanded}\\\textbf{Schaffer}}} 
 & \makecell{\shortstack{50\\\ }}  & \makecell{\shortstack{100000\\\ }} 
   & \shortstack{$2.0\cdot10^{0}$\\{\scriptsize$\pm8.5\cdot10^{-2}$}} 
   & \shortstack{$2.5\cdot10^{0}$\\{\scriptsize$\pm2.3\cdot10^{-1}$}} 
   & \shortstack{$2.1\cdot10^{0}$\\{\scriptsize$\pm7.8\cdot10^{-2}$}} 
   & \shortstack{$4.9\cdot10^{0}$\\{\scriptsize$\pm4.4\cdot10^{-1}$}} 
   & \shortstack{$\mathbf{1.4\cdot10^{0}}$\\{\scriptsize$\pm1.0\cdot10^{-1}$}} 
   & \shortstack{$3.3\cdot10^{0}$\\{\scriptsize$\pm4.2\cdot10^{-1}$}} \\
\cline{2-9}
 & \makecell{\shortstack{100\\\ }} & \makecell{\shortstack{200000\\\ }} 
   & \shortstack{$3.6\cdot10^{0}$\\{\scriptsize$\pm2.8\cdot10^{-1}$}} 
   & \shortstack{$4.1\cdot10^{0}$\\{\scriptsize$\pm7.7\cdot10^{-2}$}} 
   & \shortstack{$3.3\cdot10^{0}$\\{\scriptsize$\pm2.1\cdot10^{-1}$}} 
   & \shortstack{$5.4\cdot10^{0}$\\{\scriptsize$\pm2.9\cdot10^{-1}$}} 
   & \shortstack{$\mathbf{3.0\cdot10^{0}}$\\{\scriptsize$\pm1.3\cdot10^{-1}$}} 
   & \shortstack{$4.7\cdot10^{0}$\\{\scriptsize$\pm4.1\cdot10^{-1}$}} \\
\cline{2-9}
 & \makecell{\shortstack{200\\\ }} & \makecell{\shortstack{400000\\\ }} 
   & \shortstack{$4.2\cdot10^{1}$\\{\scriptsize$\pm3.8\cdot10^{-1}$}} 
   & \shortstack{$3.1\cdot10^{1}$\\{\scriptsize$\pm5.5\cdot10^{-1}$}} 
   & \shortstack{$\mathbf{1.6\cdot10^{1}}$\\{\scriptsize$\pm5.4\cdot10^{-1}$}} 
   & \shortstack{$5.1\cdot10^{1}$\\{\scriptsize$\pm1.9\cdot10^{0}$}} 
   & \shortstack{$2.8\cdot10^{1}$\\{\scriptsize$\pm6.5\cdot10^{-1}$}} 
   & \shortstack{$3.9\cdot10^{1}$\\{\scriptsize$\pm3.7\cdot10^{-1}$}} \\
\hline
\multirow{4}{*}{\centering\shortstack{\textbf{Schwefel}\\\textbf{with}\\\textbf{noise}}} 
 & \makecell{\shortstack{50\\\ }}  & \makecell{\shortstack{100000\\\ }} 
   & \shortstack{$3.8\cdot10^{4}$\\{\scriptsize$\pm1.5\cdot10^{3}$}} 
   & \shortstack{$\mathbf{8.3\cdot10^{3}}$\\{\scriptsize$\pm2.5\cdot10^{2}$}} 
   & \shortstack{$6.1\cdot10^{4}$\\{\scriptsize$\pm2.1\cdot10^{3}$}} 
   & \shortstack{$2.8\cdot10^{4}$\\{\scriptsize$\pm2.3\cdot10^{3}$}} 
   & \shortstack{$3.1\cdot10^{4}$\\{\scriptsize$\pm2.3\cdot10^{3}$}} 
   & \shortstack{$4.3\cdot10^{4}$\\{\scriptsize$\pm2.9\cdot10^{3}$}} \\
\cline{2-9}
 & \makecell{\shortstack{100\\\ }} & \makecell{\shortstack{200000\\\ }} 
   & \shortstack{$1.0\cdot10^{5}$\\{\scriptsize$\pm8.5\cdot10^{3}$}} 
   & \shortstack{$\mathbf{1.7\cdot10^{4}}$\\{\scriptsize$\pm4.9\cdot10^{2}$}} 
   & \shortstack{$1.6\cdot10^{5}$\\{\scriptsize$\pm3.8\cdot10^{3}$}} 
   & \shortstack{$9.2\cdot10^{4}$\\{\scriptsize$\pm3.2\cdot10^{3}$}} 
   & \shortstack{$1.2\cdot10^{5}$\\{\scriptsize$\pm4.6\cdot10^{3}$}} 
   & \shortstack{$1.0\cdot10^{5}$\\{\scriptsize$\pm3.0\cdot10^{3}$}} \\
\cline{2-9}
 & \makecell{\shortstack{200\\\ }} & \makecell{\shortstack{400000\\\ }} 
   & \shortstack{$6.9\cdot10^{5}$\\{\scriptsize$\pm6.4\cdot10^{3}$}} 
   & \shortstack{$\mathbf{9.8\cdot10^{4}}$\\{\scriptsize$\pm2.2\cdot10^{3}$}} 
   & \shortstack{$3.9\cdot10^{5}$\\{\scriptsize$\pm6.7\cdot10^{3}$}} 
   & \shortstack{$2.8\cdot10^{5}$\\{\scriptsize$\pm6.9\cdot10^{3}$}} 
   & \shortstack{$2.7\cdot10^{5}$\\{\scriptsize$\pm6.6\cdot10^{3}$}} 
   & \shortstack{$6.4\cdot10^{5}$\\{\scriptsize$\pm2.0\cdot10^{4}$}} \\
\hline
\multirow{2}{*}{\centering\shortstack{\textbf{Lennard}\\\textbf{-Jones}}} 
 & \makecell{\shortstack{50\\\ }}  & \makecell{\shortstack{100000\\\ }} 
   & \shortstack{$-1.8\cdot10^{2}$\\{\scriptsize$\pm9.5\cdot10^{0}$}} 
   & \shortstack{$\mathbf{-2.2\cdot10^{2}}$\\{\scriptsize$\pm6.8\cdot10^{0}$}} 
   & \shortstack{$-1.8\cdot10^{2}$\\{\scriptsize$\pm5.4\cdot10^{0}$}} 
   & \shortstack{$-1.9\cdot10^{2}$\\{\scriptsize$\pm6.0\cdot10^{0}$}} 
   & \shortstack{$-1.9\cdot10^{2}$\\{\scriptsize$\pm1.8\cdot10^{1}$}} 
   & \shortstack{$-1.8\cdot10^{2}$\\{\scriptsize$\pm1.5\cdot10^{1}$}} \\
\cline{2-9}
 & \makecell{\shortstack{100\\\ }} & \makecell{\shortstack{200000\\\ }} 
   & \shortstack{$-3.7\cdot10^{2}$\\{\scriptsize$\pm1.1\cdot10^{1}$}} 
   & \shortstack{$\mathbf{-4.9\cdot10^{2}}$\\{\scriptsize$\pm8.0\cdot10^{0}$}} 
   & \shortstack{$-4.1\cdot10^{2}$\\{\scriptsize$\pm1.3\cdot10^{1}$}} 
   & \shortstack{$-3.6\cdot10^{2}$\\{\scriptsize$\pm8.3\cdot10^{0}$}} 
   & \shortstack{$-4.0\cdot10^{2}$\\{\scriptsize$\pm2.0\cdot10^{1}$}} 
   & \shortstack{$-3.9\cdot10^{2}$\\{\scriptsize$\pm1.8\cdot10^{1}$}} \\
\hline
\end{tabular}%
}
\end{table}


\begin{table}[hbt!]
\centering
\caption{Best performing configurations per problem and dimension}
\label{tab:best_algorithms}
\resizebox{\textwidth}{!}{%
\begin{tabular}{|c|c|c|c|c|c|c|}
\hline
\makecell{\textbf{Problem/}\\\textbf{Dim.}} & \textbf{Sphere} & \textbf{Griewank} & \textbf{Rastrigin} & \makecell{\textbf{Expanded}\\\textbf{Schaffer}} & \makecell{\textbf{Schwefel}\\\textbf{with noise}} & \makecell{\textbf{Lennard}\\\textbf{-Jones}} \\
\hline
50  & \makecell{\textcolor{Goldenrod}{\textbf{High}}\\\textcolor{Goldenrod}{\textbf{diversity}}}   & \makecell{\textcolor{red}{\textbf{Large}}\\\textcolor{red}{\textbf{society}}}    & \makecell{\textcolor{Goldenrod}{\textbf{High}}\\\textcolor{Goldenrod}{\textbf{diversity}}}   & \makecell{\textcolor{Goldenrod}{\textbf{High}}\\\textcolor{Goldenrod}{\textbf{diversity}}}   & \makecell{\textcolor{OliveGreen}{\textbf{Explo-}}\\\textcolor{OliveGreen}{\textbf{ration}}}   & \makecell{\textcolor{OliveGreen}{\textbf{Explo-}}\\\textcolor{OliveGreen}{\textbf{ration}}}   \\
\hline
100 & \makecell{\textcolor{Goldenrod}{\textbf{High}}\\\textcolor{Goldenrod}{\textbf{diversity}}}   & \makecell{\textcolor{Goldenrod}{\textbf{High}}\\\textcolor{Goldenrod}{\textbf{diversity}}} & \makecell{\textcolor{Goldenrod}{\textbf{High}}\\\textcolor{Goldenrod}{\textbf{diversity}}}   & \makecell{\textcolor{Goldenrod}{\textbf{High}}\\\textcolor{Goldenrod}{\textbf{diversity}}}   & \makecell{\textcolor{OliveGreen}{\textbf{Explo-}}\\\textcolor{OliveGreen}{\textbf{ration}}}   & \makecell{\textcolor{OliveGreen}{\textbf{Explo-}}\\\textcolor{OliveGreen}{\textbf{ration}}}   \\
\hline
200 & \makecell{\textcolor{Goldenrod}{\textbf{High}}\\\textcolor{Goldenrod}{\textbf{diversity}}}   & \makecell{\textcolor{RedViolet}{\textbf{Small}}\\\textcolor{RedViolet}{\textbf{society}}}   & \makecell{\textcolor{RedViolet}{\textbf{Small}}\\\textcolor{RedViolet}{\textbf{society}}}   & \makecell{\textcolor{RedViolet}{\textbf{Small}}\\\textcolor{RedViolet}{\textbf{society}}}   & \makecell{\textcolor{OliveGreen}{\textbf{Explo-}}\\\textcolor{OliveGreen}{\textbf{ration}}}   & --  \\
\hline
\end{tabular}%
}
\end{table}




We systematically performed several statistical tests on the quantitative results obtained in this study. Initially, we applied the Shapiro–Wilk test ($\alpha = 0.05$) to assess whether the observed samples followed a normal distribution. The null hypothesis was rejected for the samples corresponding to IM and each tested configuration. Thus, we used the Kruskal–Wallis test, which revealed significant differences in cumulative distribution functions across configurations. %
\enlargethispage{1.5\baselineskip}%
Consequently, we applied Dunn’s post-hoc test with Holm-Bonferroni correction for multiple comparisons. Table~\ref{tab:dunnfail} lists configuration pairs involving the reference baseline IM that did not yield statistically significant adjusted p‑values ($\alpha = 0.01$).

\begin{table}[h]
    \centering
    \caption{Dunn test p-values for algorithm pairs (Holm-Bonferroni--adjusted) with the baseline that exceeded the 0.01 threshold and are considered not significantly different.}
    \label{tab:dunnfail}
    \begin{tabular}{|l|c|l|c|}
        \hline
        \textbf{Problem} & \textbf{Dim.} & \textbf{Pair} & \textbf{p-value} \\
        \hline
        Sphere & 50 & Strong leadership vs Island model & 0.2220 \\
        Sphere & 50 & Small society vs Island model & 0.2220 \\
        Griewank & 50 & Small society vs Island model & 0.5741 \\
        Griewank & 50 & Large society vs Island model & 0.5741 \\
        Griewank & 200 & Exploration vs Island model & 0.2897 \\
        Rastrigin & 50 & High diversity vs Island model & 0.2298 \\
        Rastrigin & 100 & Small society vs Island model & 0.1165 \\
        Expanded Shaffer & 100 & Exploration vs Island model & 0.1003 \\
        \hline
    \end{tabular}
\end{table}




\section{Conclusions} \label{sec:summary}

This study introduce the Trust-Based Optimization (TBO) algorithm, an extension of the island model that involves trust and reputation mechanisms to enable adaptive information exchange.

Experimental results demonstrate that TBO can improve convergence speed and maintain population diversity more effectively in specific configurations. The \textit{Exploration} and \textit{High Diversity} configurations performed well across multiple problem types, while \text{Small society} configuration proved effective in higher-dimensional problems and \textit{Strong Leadership} facilitated rapid early-stage optimization. However, no single configuration was universally superior, as performance varied depending on the problem. The Schwefel with Noise and Lennard-Jones Minimum Energy Cluster problems posed particular challenges, suggesting that TBO's effectiveness may be limited in highly deceptive or noisy landscapes.


Overall, TBO successfully integrates socio-cognitive concepts into evolutionary optimization, offering a flexible and adaptive approach that enhances algorithmic performance. Future work may explore the impact of heterogeneous agent configurations within a single TBO system or hybrid approaches combining TBO with game theory, reinforcement learning, dynamic adaptation of interaction parameters, or its application to real-world optimization problems.

\begin{credits}
\subsubsection{\ackname} The research presented in this paper has been financially supported by: Polish National Science Center Grant no. 2019/35/O/ST6/00570 ``Socio-cognitive inspirations in classic metaheuristics'';  Polish Ministry of Science and Higher Education funds assigned to AGH University of Science and Technology, program „Excellence initiative – research university" for the AGH University of Krakow. ARTIQ project – Polish National Science Center:DEC-2021/01/2/ST6/00004, Polish National Center for Research and Development, DWP/ARTIQI/426/2023 (MKD)

\subsubsection{\discintname}
The authors have no competing interests to declare that are
relevant to the content of this article. 
\end{credits}
%
%
%

\enlargethispage{2.5\baselineskip}
\bibliographystyle{splncs04}
\bibliography{sc_trust}

\end{document}